\documentclass{article}

\usepackage{graphicx} %插入图片的宏包
\usepackage{float} %设置图片浮动位置的宏包
\usepackage{subfigure} %插入多图时用子图显示的宏包

\usepackage{threeparttable}

\usepackage[preprint]{neurips_2022}

\usepackage[utf8]{inputenc} % allow utf-8 input
\usepackage[T1]{fontenc}    % use 8-bit T1 fonts
\usepackage{hyperref}       % hyperlinks
\usepackage{url}            % simple URL typesetting
\usepackage{booktabs}       % professional-quality tables
\usepackage{amsfonts}       % blackboard math symbols
\usepackage{nicefrac}       % compact symbols for 1/2, etc.
\usepackage{microtype}      % microtypography
\usepackage{xcolor}         % colors

\title{An Adaptive Contrastive Learning Model for Spike Sorting}

\author{
  Lang Qian\thanks{Lang Qian and Shengjie Zheng contributed equally to this work} ,  Cheng Yang\\
  Tsinghua Shenzhen International Graduate School \\
  Tsinghua University \\
  Shenzhen, China \\
  \texttt{ql20@mails.tsinghua.edu.cn, yang.cheng@sz.tsinghua.edu.cn} \\
  \And
  Shengjie Zheng \\
  Shenzhen Institute of Advanced Technology, Chinese Academy of Sciences \\
  University of Chinese Academy of Sciences \\
  Shenzhen, China \\
  \texttt{sj.zheng@siat.ac.cn} \\
  \And
  Chunshan Deng, Xiaojian Li \thanks{Cheng Yang and Xiaojian Li are co-corresponding authors}\\
  Brain Cognition and Brain Disease Institute (BCBDI) \\
  Shenzhen Institute of Advanced Technology, Chinese Academy of Sciences \\
  Shenzhen, China \\
  \texttt{cs.deng@siat.ac.cn, xj.li@siat.ac.cn} \\
}

\begin{document}

\maketitle

\begin{abstract}
Brain-computer interfaces (BCIs), is ways for electronic devices to communicate directly with the brain. For most medical-type brain-computer interface tasks, the activity of multiple units of neurons or local field potentials is sufficient for decoding. But for BCIs used in neuroscience research, it is important to separate out the activity of individual neurons. With the development of large-scale silicon technology and the increasing number of probe channels, artificially interpreting and labeling spikes is becoming increasingly impractical. In this paper, we propose a novel modeling framework: Adaptive Contrastive Learning Model that learns representations from spikes through contrastive learning based on the maximizing mutual information loss function as a theoretical basis. Based on the fact that data with similar features share the same labels whether they are multi-classified or binary-classified. With this theoretical support, we simplify the multi-classification problem into multiple binary-classification, improving both the accuracy and the runtime efficiency. Moreover, we also introduce a series of enhancements for the spikes, while solving the problem that the classification effect is affected because of the overlapping spikes.
\end{abstract}

\section{Introduction}

Brain-computer interfaces (BCIs), are ways for electronic devices to communicate directly with the brain. The most common brain-computer interface applications are in medical care, typically in the form of collecting neural signals from a subject's brain to help the subject perform tasks such as spelling words\cite{handwriting}, controlling a computer cursor\cite{cursor}, controlling a mechanical prosthesis\cite{even2020power}, or paralyzed muscles, and interacting with the outside world. On the other hand, in the field of neuroscience research, brain-computer interfaces are also powerful and unique tools for studying brain function. So far, the most reliable way to record neural activity is to put microelectrodes into the brain to record local electrical activity, from which we can observe the action potential. We can assign it to the correct neuron by detecting the channel where the action potential appears, the amplitude and waveform of spike, etc. This process of neural information processing is called spike sorting\cite{gibson2008comparison}\cite{li2019supervised}. However, microelectrodes are still affected by different factors that make it hard to interpret and challenge to assign this recorded waveform to the correct neuron, such as microelectrode floating, spike overlapping, etc.

Nowadays, big data and high performance computing power have allowed deep learning technology to make rapid development in applications, especially supervised learning.Although supervised learning has been very successful on CV and NLP, we cannot directly apply supervised learning to spike sorting. Even if the model gets high accuracy, the model cannot transfer effectively because same neurons from different subjects could fire different spike waveforms.  Moreover, the artificial labeling of spike is not only very troublesome in processing, but also requires professionals to spend a lot of time\cite{li2019supervised}\cite{spikedeep}. At the same time, the use of supervised learning can cause overfitting of the neural network when spike overlapping is frequently encountered. Based on this, there is a slow shift towards unsupervised but non-contrastive learning\cite{AE}\cite{seong} in dealing with spike sorting, which does not use labels, but the purpose of such models is only to reconstruct the input and 
extract the features of the middle layer for clustering, without really extracting the latent representation of spikes fired by different neurons effectively. 

With the rapid development of contrast learning, the ability of contrast learning to extract the latent representation has caught up with supervised learning in some aspects. Based on this situation, we propose a new framework based on contrast learning that (1) simplifies a multi-classification task into multiple binary-classification tasks, (2) proposes a series of augmented approaches for spikes, and (3) introduces contrastive learning into spike sorting. Our approach concerns extracting features of spike data from all channels at the same time interval, rather than considering a single channel spike data simultaneously. Because we believe that the multi-channel spike data carries more valid information and can allow us to achieve better classification effect of spikes.

In order to better achieve the latent representation of spikes, we use a method named invariant information clustering which maximizes the mutual information of raw data and augmented data\cite{IIC}. In addition, we used a series of augmentation methods of the data, with the aim of generating data sharing similar semantics and  the same labels as the original data.  In the past two years, contrastive learning has been used both in CV\cite{SimCLR} or in NLP\cite{yan2021consert}\cite{SimCSE} have had great success on it. But for spike data augmentation, which is not the same as image or text augmentation, we propose a series of spike augmentation methods based on the theoretical basis of electrophysiology. We also tested the accuracy of our model on a simulated dataset generated by kilosort\cite{KiloSort} and in-vivo data from rat hippocampus\cite{mizuseki2013multiple}.

\section{Method}
\label{gen_inst}

\subsection{Data Preprocessing}

\begin{figure}[htbp] %H为当前位置，!htb为忽略美学标准，htbp为浮动图形
\centering %图片居中
\includegraphics[scale=0.3]{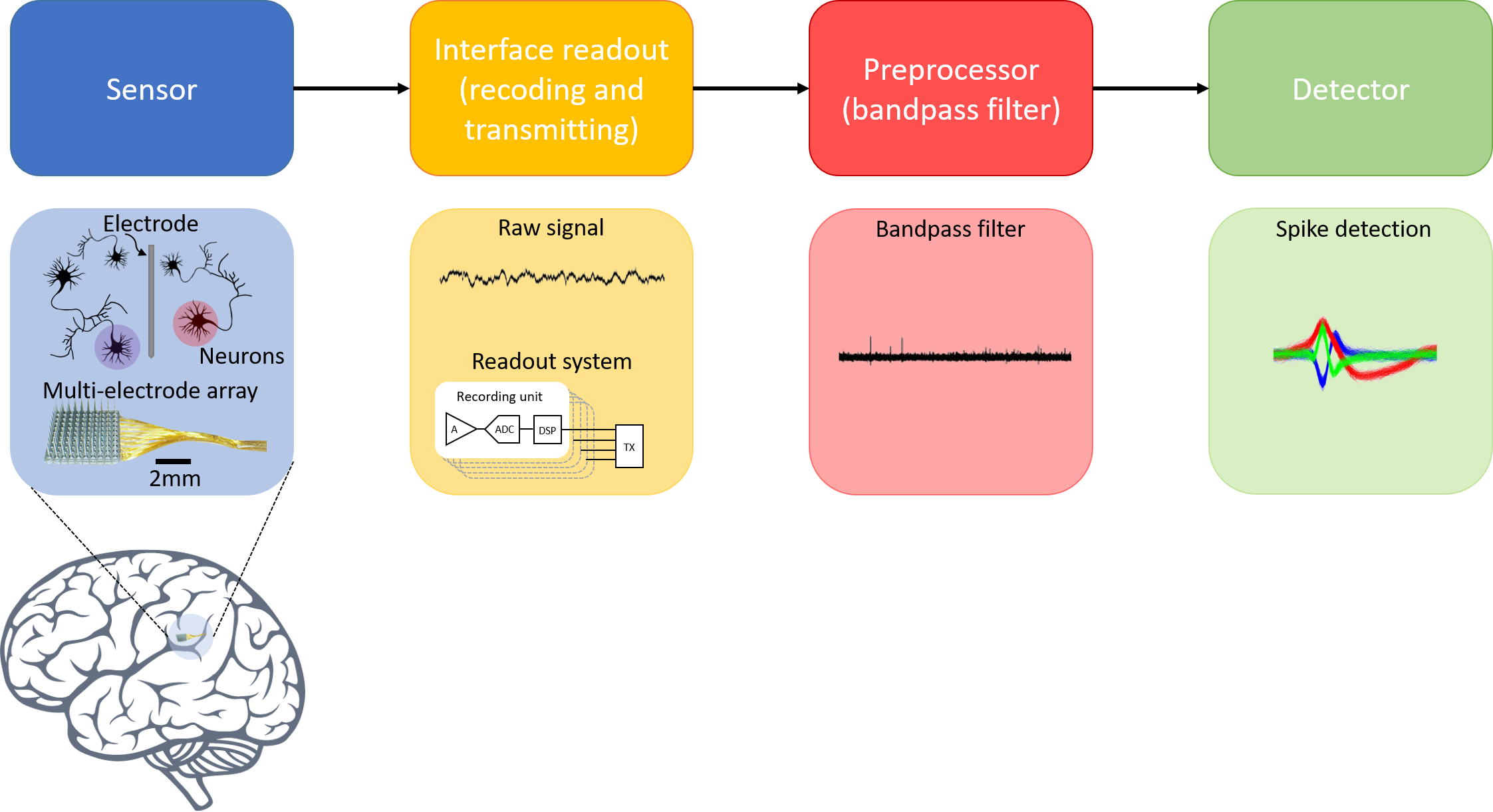} %插入图片，[]中设置图片大小，{}中是图片文件名
\caption{The neural information is recorded using an array of electrodes, each measuring nearby neural activity. The raw neural signal is transformed into a digital signal by the signal acquisition system. The signals can be processed by band-pass filters and spike detecting by detector. The image modified from \cite{zheng2022spiking}.} %最终文档中希望显示的图片标题
\label{Fig.main0} %用于文内引用的标签
\end{figure}

Step 1: The signal recordings obtained through microelectrodes are not directly usable due to Local Field Potential (LFP) and high frequency noise. We need to band-pass filter the raw data from 300Hz to 3000Hz, the data in the range from 1Hz to 100Hz, which is usually LFP, and the data in the range from 300Hz to 3000Hz, which is the spike we use normally, the data at higher frequencies are usually high frequency noise \cite{past}, As shown in Fig.\ref{Fig.main0}.

Step 2: When we get the recorded bandpass filtered signal, we need to locate the spikes, there are many methods for example, Absolute Value, Nonlinear Energy Operator, Stationary Wavelet Transform Product \cite{gibson2008comparison}, the task of locating the spikes is not the focus of this study.

\subsection{Data Augmentation}

The key element of contrast learning is to construct augmented data with the same semantic meaning as the original sample. We can use contrast learning to find the maximum similarity in similar samples and the maximum difference in different samples and based on that, we can extract the latent representation in the samples. For example, we can recolor the image or scale the audio amplitude, but we can still clearly perceive that the augmented data share the same semantics as the original data \cite{SimSEQ}. However, the augmentation methods of spike data is not exactly the same as the augmentation of image data, text data, and audio data.  To solve the problem of spike data augmentation, we want to propose a system on multi-channel spike data augmentation that can guarantee that the augmented sample and the original sample have the same semantics between them both.

\begin{figure}[htbp] %H为当前位置，!htb为忽略美学标准，htbp为浮动图形
\centering %图片居中
\includegraphics[scale=0.2]{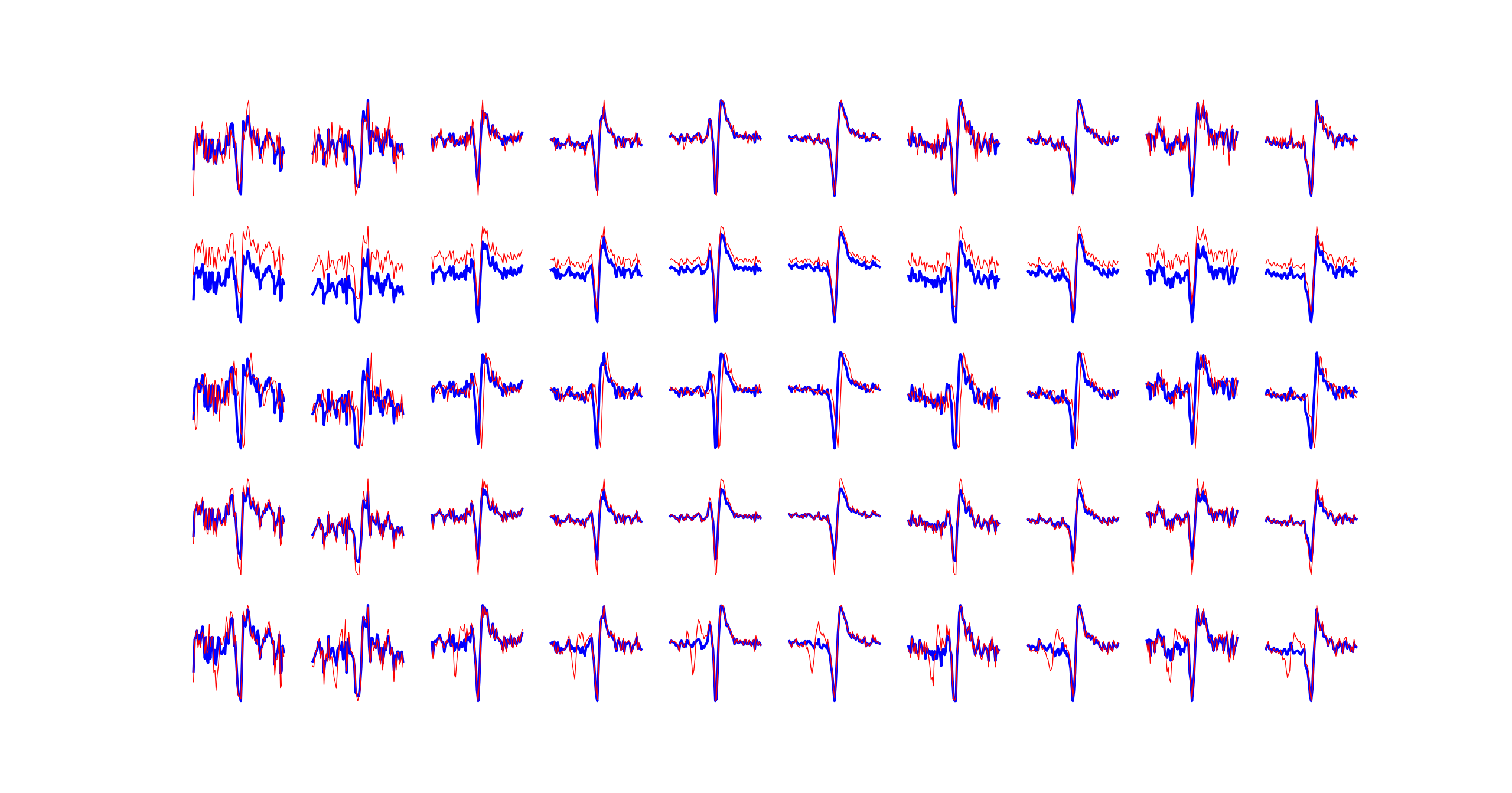} %插入图片，[]中设置图片大小，{}中是图片文件名
\caption{Five enhancements of spike data are described, from top to bottom, as random noise, DC shifting, Horizontal shifting, Amplitude scaling, Spike overlapping. } %最终文档中希望显示的图片标题
\label{Fig.main1} %用于文内引用的标签
\end{figure}

We uphold the idea that problem solving starts from the problem itself. We augmented the spikes fired by neurons, and we need to try to simulate the different spike data fired by neurons in the brain due to various possible influences. This is because the spike fired by the same neuron at different moments is not exactly the same, but they share similar semantics. After observing the data fired by neurons and consulting with relevant biological background experts, we propose five multi-channel spike data augmentation methods, as shown in Fig. \ref{Fig.main1}.(1) Random noise, (2) DC shifting (3) Horizontal shifting (left or right),  (4) Amplitude scaling, and (5) Spike overlapping.

This paragraph focused on the details of these five data augmentation methods. the operation is performed on normalized data, with 10 channels and 81 sampling points.(1) Random noise: the distribution of random noise values is a random normal distribution with a variance of 0.1. (2) DC shifting: obeying a uniform distribution of - 0.1 to 0.1. (3) Horizontal shifting: Obeying a uniform distribution of - 20 to 20 and the value is an integer. (4) Amplitude scaling: Following the uniform distribution of 0.9 to 1.1. (5) Spike overlapping: randomly selecting a data (including raw data), performing horizontal offset, then randomly reducing the amplitude and summing it with the raw data.

\subsection{Model Architecture}

\begin{figure}[H] %H为当前位置，!htb为忽略美学标准，htbp为浮动图形
\centering %图片居中
\includegraphics[width=0.7\textwidth]{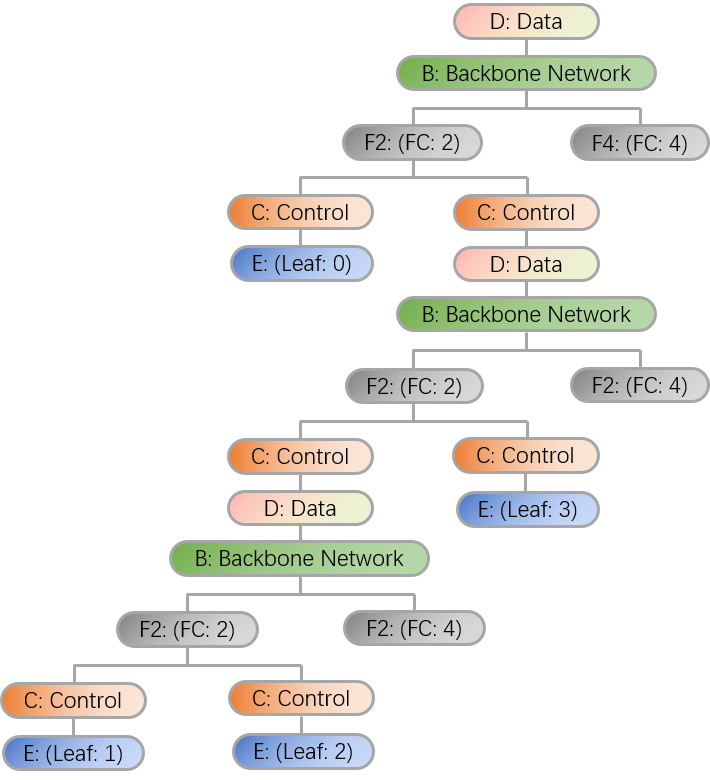} %插入图片，[]中设置图片大小，{}中是图片文件名
\caption{Describing the ACCM model architecture} %最终文档中希望显示的图片标题
\label{Fig.main2} %用于文内引用的标签
\end{figure}

Classifying spike data is not the same as classifying pictures in the traditional way, because we cannot determine the number of spike classifications, we cannot treat classifying spike data as a multi-classification task. In this case, we propose a new spike data classification model: an adaptive contrast classification model, referred to as ACCM, as shown in Fig. \ref{Fig.main2}. 

In this model, to give users more choices, we design three encoder architectures for Backbone Network, as shown in Fig. \ref{Fig.main2}: 

(1) A transformer encoder is abbreviated as ACCM-T, as shown in Fig.4(a). This model not only focuses on the peak and waveform of spike data, but also pays more attention to the channel-to-channel relationship. Based on this situation, we chose the Transformer encoder block because of its unique Multi-Head attention mechanism. Also, to focus more on the channel information, this model still includes the position encoding operation, which is also suitable for large batches of data.

(2) A 1D-CNN encoder is abbreviated as ACCM-C, as shown in Fig.4(b). With aim of extracting the different latent features of the current channel spike, the data have three 1D-convolution operations. The kernel size of three 1D-convolution operations is all 1. Meanwhile, the output dimensions are 5,10,15 repectively. The reason for this design is to pay more attention to the characteristics of the spike itself in terms of waveform, amplitude and other relevant factors. After three 1D-convolution operations, we performed the concatenate and flatten operations respectively. We also abandoned the use of L2-Regularization with considering that spike data is very sensitive to amplitude.  Finally, we reduced the dimensionality of the data by performing several fully connected operations. Meanwhile, we abandon the use of 2D-CNN in this model, because the relationship between the microelectrode channels is not really linear, but more like a graph structure. This model encoder is more suitable for spike data that needs to be sensitive to peaks and patterns.

(3) A recurrent neural network encoder is abbreviated as ACCM-R, as shown in Fig.4(c). This model is more concerned with channel-to-channel relationships, when the number of channels of neural electrodes reaches hundreds or thousands, not all channels can detect spike signals, and the neural information contained in multiple channels is sufficient to meet the requirements of spike sorting. This model encoder is more suitable for channel-sensitive spike data.

At the same time, when we train the model, in order to avoid falling into the local optimal point, we iteratively trained the two-class and four-class classifiers, which can help us find the global optimal point. We set up a controller under each binary-classification branch to decide whether the classified data should be classified again.

\begin{figure}[H]
\centering
\includegraphics[scale=0.5]{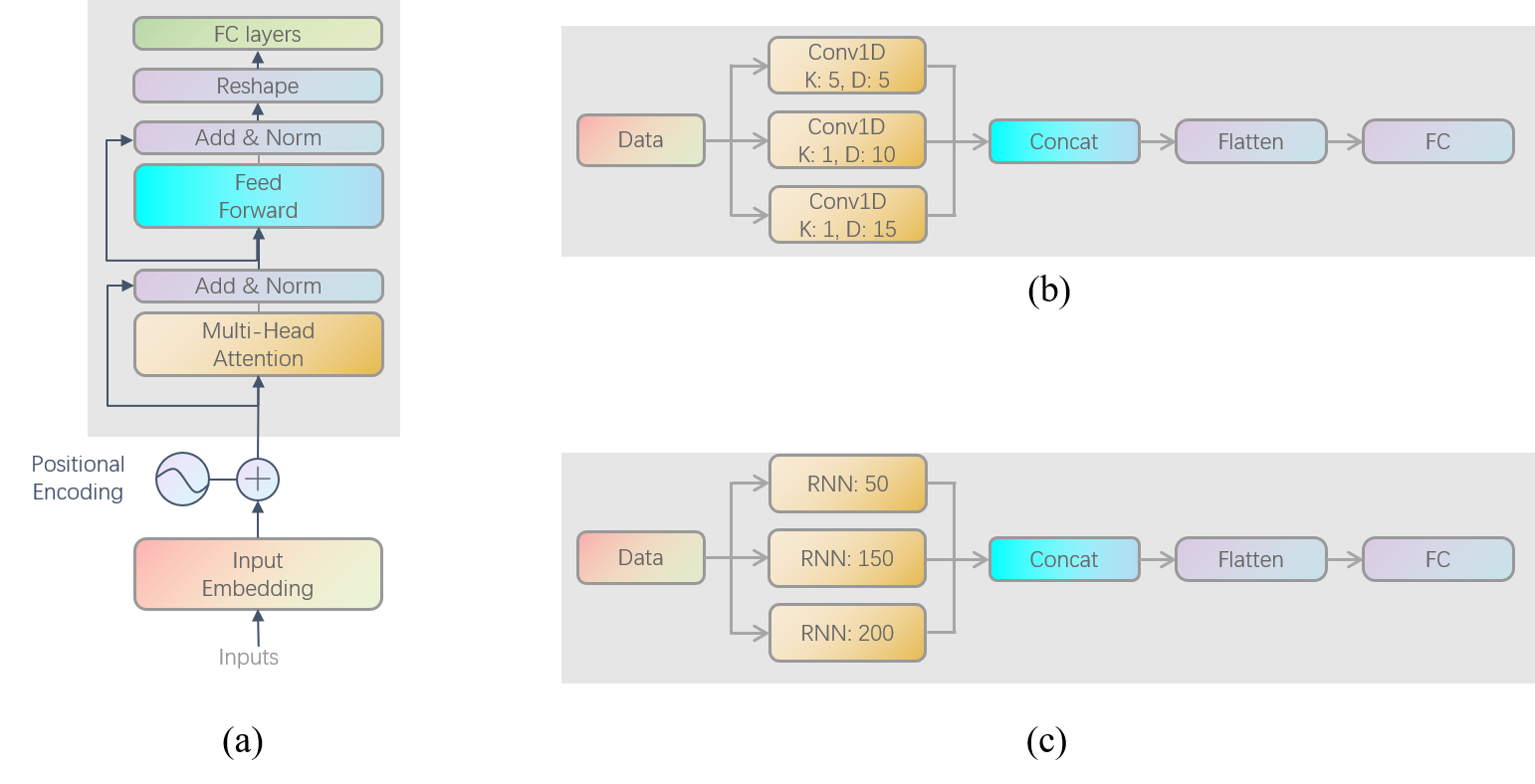}
\caption{Encoder architectures: (a) Transformer encoder block, (b) 1D-CNN encoder block, (c) RNN encoder block.}
\label{figureblock}
\end{figure}

\subsection{Contrastive loss}

In this model, we choosed to maximize the mutual information for contrastive loss\cite{IIC}. Because we are dealing with a binary-classification task, so the joint probability distribution was given by the 2x2 matrix $\mathrm{M}$, as shown in Equation \ref{Equa1}.

\begin{equation}
\mathrm{M}=\frac{1}{n} \sum_{i=1}^{n} \Phi\left(\mathbf{x}_{i}\right) \cdot \Phi\left(\mathbf{x}_{i}^{\prime}\right)^{\mathrm{T}}
\label{Equa1}
\end{equation}

The output $\Phi(\mathbf{x}) \in[0,1]^{2}$ can be interpreted as the distribution of a discrete random variable over $2$ classes. As we generally consider symmetric problems, we usually redefine $\mathrm{M}$ as $\left(\mathrm{M}+\mathrm{M}^{\mathrm{T}}\right) / 2$.

\begin{equation}
I(\mathrm{M})=\sum_{c=1}^{2} \sum_{c^{\prime}=1}^{2} \mathrm{M}_{c c^{\prime}} \cdot \ln \frac{\mathrm{M}_{c c^{\prime}}}{\mathrm{M}_{c} \cdot \mathrm{M}_{c}^{\prime}}
\label{Equa2}
\end{equation}

where maximizing $I(\mathrm{M})$ trades-off minimizing the difference between the raw data and the augmented data, as shown in Equation \ref{Equa2}.

\subsection{Dataset}

\textbf{In-vivo dataset:} We used the hc3 dataset sourced from CRCNS\cite{mizuseki2013multiple}, the data set contains recordings made from multiple hippocampal areas in Long-Evans rats. The raw (broadband) data was recorded at 20KHz, and we selected a segment of data with 8 channels, each with 32 sampling points, and a total of three neurons firing spikes. Total number of spike data is 6000, each neuron fired a total of 2,000 times.

\textbf{Simulated data:} The simulated dataset we used was derived from kilosort, having 10 channels, each with 81 sampling points, and a total of eleven neurons firing spikes. Total number of spike data is 22000, each neuron fired a total of 2,000 times\cite{KiloSort}.

\section{Experiment}
\label{Exp}

In this section, we compared our results with one unsupervised model AE \cite{AE} which reconstructed the model own input  and two contrastive learning models, SimCLR \cite{SimCLR}, SimCSE \cite{SimCSE}. The initialization weights of the model parameters and the selection of the optimizer as well as the size of the learning rate are also presented.

\subsection{Model Training}

To train our model, we experimented with three architectures Fig. \ref{figureblock}. In each backbone network, the choice of batch sizes with training data is not exactly the same. The epochs of each backbone network node are determined by the number of training data, and usually fluctuate between 15 and 60. After several model training sessions, we found that initializing the model parameters to a random normal distribution with variance of 0.1 allowed the model to converge faster. At the same time, we use the Adam optimizer instead of the SGD optimizer in order to avoid getting trapped in the local optimal point, and the learning rate was initialized to 0.005. We performed the experiments on one GTX 2080ti GPU.

\section{Results}
\label{others}

We tested three different versions of the model, ACCM-R, ACCM-C, and ACCM-T, and obtained the results of these three versions in terms of accuracy and running time, respectively, for comparison with an unsupervised model AE \cite{AE}, two contrastive learning models SimCLR\cite{SimCLR} and SimCSE\cite{SimCSE}.

% \usepackage{booktabs}

\begin{table}[H]
\centering
\caption{Simulated}
\label{tab:my-table1}
\begin{tabular}{@{}llc@{}}
\toprule
Model  & ACC    & \multicolumn{1}{l}{Running Time} \\ \midrule
AE     & 83.2\% & 6min13s              \\
\hline
SimCLR & 87.8\% & 8min17s                          \\
SimCSE & 61.1\% & 3min3s                           \\
\hline
ACCM-R & 95.3\% & 21min3s                          \\
ACCM-C & 96.5\% & 5min41s                          \\
ACCM-T & 98.4\% & 11min4s                          \\ \bottomrule
\end{tabular}
\end{table}

% \usepackage{booktabs}
\begin{table}[H]
\centering
\caption{In-vivo}
\label{tab:my-table2}
\begin{tabular}{@{}llc@{}}
\toprule
Model  & ACC    & \multicolumn{1}{l}{Running Time} \\ \midrule
AE     & 71.2\% & 1min24s                          \\
\hline
SimCLR & 97.9\% & 1min46s                          \\
SimCSE & 82.1\% & 1min21s                           \\
\hline
ACCM-R & 98.7\% & 1min9s                          \\
ACCM-C & 99.9\% & 1min6s                          \\
ACCM-T & 99.9\% & 58s                          \\ \bottomrule
\end{tabular}

\end{table}

The above table \ref{tab:my-table1} and table \ref{tab:my-table2} shows the results of our experiments, the second column represents the accuracy rate, and the third column is the model running time to reach the highest accuracy rate.
Our model is provided in three different versions for different situations. If we need fast running time and relatively high accuracy, we can choose ACCM-C version. For high efficiency, we can choose ACCM-T. Also, ACCM-T achieves very high accuracy on simulated data.

Through the above tables,we observed that the accuracy of model SimCSE is much less than that of model SimCLR for both in-vivo and simulated datasets. Model SimCSE augmented spike data only by dropout, while model SimCLR augmented spike data through our model's data augmentation methods. This can illustrate the superiority of our data augmentation methods. Meanwhile, we observe that model SimCLR is much more accurate when the number of clusters is relatively small than when the number of clusters is relatively large. This illustrated that in spike sorting, multiple binary-classification tasks could be easier to get a higher accuracy rate than  the multi-classification task. Again, we concluded from the above two tables that our model could obtain high accuracy in both in-vivo and simulated datasets, which is a good indication of the superiority of our model in handling spike sorting.

\section{Conclusion}
\label{Conclusion}

In the field of neuroscience research, brain-computer interfaces provide a way to study the direct causal relationship between small populations of neurons and specific external outputs. Identification of the spiking waveform of each neuron is the basis for distinguishing the activity of individual neurons in a small population of neurons. In this paper we introduced ACCM, a self-supervised framework for learning representation for spike data. To be able to accomplish this work, we proposed five ways of augmenting the spike data. We proposed three model variants, ACCM-C, ACCM-R, and ACCM-T, because different professionals have different preferences for accuracy and runtime when classifying spikes, and for the first time we introduced contrastive learning into spike sorting, achieving state-of-the-art accuracy without the need for electrode maps. And, since we cannot determine the number of clusters while performing spike sorting, we also proposed an adaptive classification algorithm to transform the multi-classification task into multiple binary-classification tasks.

{
\small
\bibliographystyle{plainnat}
\bibliography{neurips_2022}
}

\end{document}